\documentclass[final,3p,times,twocolumn]{elsarticle}
\usepackage{amssymb}
\usepackage{url}

\usepackage[final]{listofsymbols}
\usepackage{mathtools}
\usepackage{textcomp}
\usepackage{multirow}
\usepackage{graphicx}
\journal{}

\begin{document}

\begin{frontmatter}



\title{Multi-task Explainable Skin Lesion Classification}


\author{Mahapara Khurshid}
\author{Mayank Vatsa}
\author{Richa Singh}
\affiliation{Indian Institute of Technology, Jodhpur, India}

\begin{abstract}
Skin cancer is one of the deadliest diseases and has a high mortality rate if left untreated. The diagnosis generally starts with visual screening and is followed by a biopsy or histopathological examination. Early detection can aid in lowering mortality rates. Visual screening can be limited by the experience of the doctor. Due to the long tail distribution of dermatological datasets and significant intra-variability between classes, automatic classification utilizing computer-aided methods becomes challenging. In this work, we propose a multitask few-shot-based approach for skin lesions that generalizes well with few labelled data to address the small sample space challenge. The proposed approach comprises a fusion of a segmentation network that acts as an attention module and classification network. The output of the segmentation network helps to focus on the most discriminatory features while making a decision by the classification network. To further enhance the classification performance, we have combined segmentation and classification loss in a weighted manner. We have also included the visualization results that explain the decisions made by the algorithm. Three dermatological datasets are used to evaluate the proposed method thoroughly. We also conducted cross-database experiments to ensure that the proposed approach is generalizable across similar datasets. Experimental results demonstrate the efficacy of the proposed work.

\end{abstract}


% \begin{highlights}
% \end{highlights}


\end{frontmatter}


\section{Introduction}
\label{}
{S}{kin} diseases is a common human illness that affects between 30 to 70\% of people and can worsen if left untreated. These diseases can be caused by a number of factors, including microscopic bacteria, a fungus that thrives on the skin, some kind of allergies or pigmentation \cite{aldhyani2022multi}. These cause lesions on the skin. The skin lesions can refer to any imperfection or defect located either above or below the skin surface. These lesions have been divided into two categories - benign and malignant. Benign lesions correspond to non-threatening skin tumours such as cysts or moles; malignant tumours are cancerous lesions, including melanoma, squamous cell carcinoma, and basal cell carcinoma, among others \cite{benyahia2022multi}. Melanoma originates from the melanocyte cells having a mole-like appearance and is often black or brown in colour \cite{khan2019construction}. Despite the fact that melanoma constitutes only 7\% among others, it has a mortality rate of 75\%, making it the deadly type of skin cancer \cite{afza2022multiclass}. According to WHO \cite{currentstats}, there were around 324,000 "melanoma of the skin" cases globally in 2020. Furthermore, roughly 106,110 additional cases and 7,180 skin cancer deaths are estimated in the United States alone in 2021, according to \cite{siegel2021cancer}. Early detection of melanoma can prevent metastases and, as a result, mortality in the majority of cases. 

An experienced dermatologist diagnoses skin cancer visually, followed by various clinical tests such as biopsy or histopathological examination. Generally, the doctors use dermoscopy, a non-invasive procedure, that magnifies the region at a high resolution and help them to clearly identify the minute spots and other details of the lesions. However, it is limited by human vision, perception and sensitivity. It requires significant training and experience in addition to the time-consuming procedure. This triggered the research community to develop precise automated approaches to assist doctors in early screening. This research aims to propose a non-invasive computer-aided screening method to help classify skin lesions and aid in the visual screening process.

\begin{figure}[!tb]

  \centering
  \centerline{\includegraphics[width=3.2in]{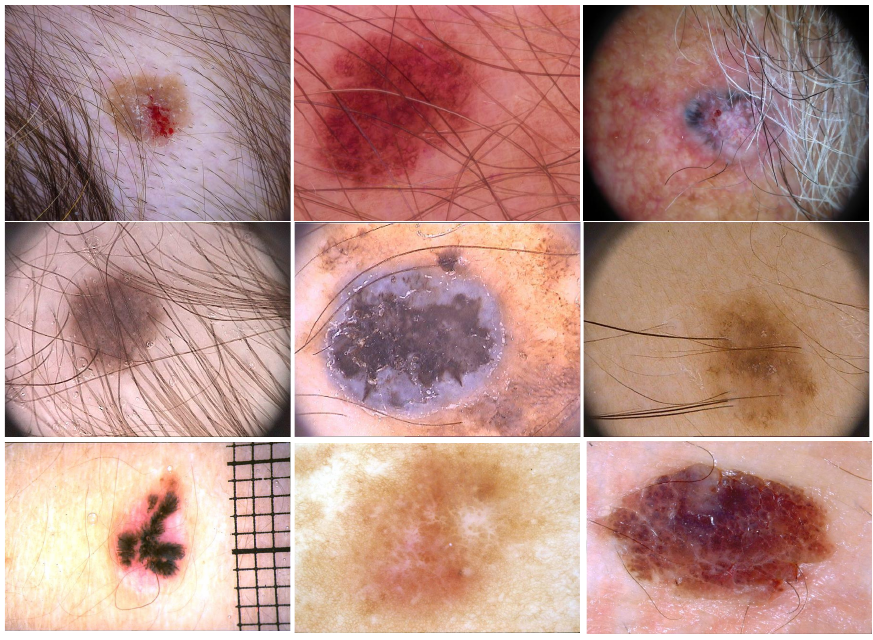}}
%  \vspace{2.0cm}
  \caption{Showcasing the skin samples from HAM10000 (top row), PH2 (middle row) and Derm7pt (bottom row). This figure highlights the need for segmentation in skin lesion classification, as there are artifacts that need to be removed before classifying the image.}
  \label{fig:intro}
  \end{figure}

Classification of skin lesion images has a relatively extensive literature. Some early methods use handcrafted features \cite{ballerini2013color,cheng2008skin,barata2013two, tommasi2006melanoma} such as texture, colour, and shape. However, these methods have limitations due to high visual similarity, intra-class variability, and the presence of various artefacts. Furthermore, these methods were dataset-specific and did not generalize well across datasets. In addition to the boundary between the lesion and the healthy skin, as shown in Fig. \ref{fig:intro}, artefacts such as hair, reference scales and low contrast are examples that might affect classification performance. This leads to the need to extract the region of interest (ROI) before performing classification. It led to the development of approaches that are dependable, resilient, and generalizable.

% Although deep learning has shown promising results in various fields, including medicine, this success is mainly attributed to the availability of large annotated datasets. 
In the field of medicine, deep learning has had a lot of success \cite{wu2019deep,li2020difficulty,tripathi2022mtcd, ngai2022emotion,yin2022df}. These approaches have shown good performance \cite{zhu2021fusing}, can automatically learn relevant features \cite{liu2017multi}, are capable of handling large and complex datasets and are highly adaptable.
% The availability of large scale labelled datasets is mainly responsible for this success.
Despite advancements in deep learning algorithms, there are some challenges that need research focus. The first challenge is the lack of large annotated datasets, particularly for rare diseases. Using deep learning on small datasets will lead to overfitting of the model. Hence, learning a reliable and robust model with a small number of samples is another challenge. Existing techniques are ineffective in dealing with such scenarios where retraining on new data is required. Due to the small number of available samples, existing techniques may not be able to generalize effectively, resulting in lower performance.

% The research community is experimenting with few-shot learning based approaches to address data scarcity and the availability of a large number of annotated samples. 
The research community is focusing on developing approaches to overcome the aforementioned challenges. One line of research is the few-shot learning-based approaches. These approaches can be meta-learning-based \cite{finn2017model}, metric-based \cite{snell2017prototypical,sung2018learning}, or optimization-based \cite{ravi2016optimization}. These approaches are gaining popularity because the model generalizes well to fewer data samples.
Despite achieving success in a variety of tasks, an automated medical diagnosis system ought to be transparent, comprehensible, and explainable. One of the most desired properties is explainability \cite{samek2017explainable}. Explainability is referred to as the process of examining the decisions that the system has made. It should be able to justify the reasoning behind any decision. The end-user or health professional should be able to retrace the decisions that inculcate trust among them. Additionally, it is important to demonstrate features in  such a way that is fully understandable to non-deep learning users, such as medical professionals. 

Continuing in the path of dealing with low data regimes and introducing explainability, this work aims to present a multitask explainable few-shot learning-based approach by fusing segmentation and classification tasks for skin lesions. In this work, we use prototypical networks, a metric-based few-shot learning approach. This approach is more straightforward, robust and computationally efficient as compared to other approaches. To extract the ROI from the input image, we have added a segmentation step in our proposed approach that will extract the ROI from the input image and enable the classification model to focus on only that region.  We have also added visual descriptions of the decisions made by the system that aid in comprehending the regions that were focused on while making any decision. 

To summarise, we propose a multitask few-shot learning technique for skin lesion classification that fuses segmentation and classification tasks in a unified manner. The following are the key contributions of this paper:
\begin{itemize}
    \item We propose a prototypical few-shot learning-based end-to-end network for the skin lesions
    \item The proposed framework combines the segmentation and classification in such a way that the artefacts are excluded before classifying the input image
    \item We evaluate the performance of the proposed method in comparison to the Cosine (Cos) and Euclidean (Euc) distances
    \item Cross-database experiments are conducted utilising the HAM10000, Derm7pt and PH2 datasets to illustrate the generalizability of the proposed approach.
    \item Explainability of the proposed approach is illustrated using GradCAM results.
\end{itemize}

\section{Related Work}
 This section provides a brief overview of approaches developed for segmenting and classifying skin lesion images.

\subsection{Skin Lesion Segmentation}
Jafari et al. \cite{jafari2016skin} have proposed using a guided filter as a pre-processing filter and extracted local and global patches based on fixed window size to classify the pixel as a lesion or normal. Xue et al. \cite{xue2018adversarial} have proposed an adversarial-based skin lesion segmentation network. In this work, the authors have proposed a multi-scale L1 loss function that learns global and local features by capturing long- and short-range pixel relationships. Wei et al. \cite{wei2019attention} conducted a similar research in which the authors utilised an attention module that automatically suppresses irrelevant features and focuses just on lesion features. In addition, the authors propose a new loss function that combines the Jaccard distance loss and the adversarial feature mapping loss. Experiments show that their proposed method produces precise masks with sharp boundaries. A GAN-based strategy for skin lesion segmentation has also been proposed by Tu et al. \cite{tu2019dense}. The authors employed a network with a Dense-residual block to help in feature propagation and reuse and multi-scale feature mapping from several layers. Bi et al. \cite{bi2019improving} have developed a stacked adversarial learning technique for segmenting skin lesions. The authors iteratively learned class-specific features to maximise feature diversity and added them to the FCN training data. The results of the experiments show that this method enhanced segmentation accuracy.

Mishra et al. \cite{mishra2017deep} have proposed an algorithm for skin lesion segmentation using the concept of U-Net architecture with fewer layers. The authors have compared their results with otsu's thresholding method and concluded that deep learning gives better results. SkinNet, a modified version of U-Net, is proposed by Vesal et al. \cite{vesal2018skinnet}. The architecture employed dilated and densely block convolutions during training to incorporate multiscale and global information from the input image. The results show that their proposed method can handle the poor contrast between the lesion and healthy tissue. Hasan et al. \cite{hasan2020dsnet} proposed the DSNet, a semantic segmentation network for skin lesions. To reduce the number of parameters, the authors used dense blocks in the encoder and depth-wise separable convolutions in the decoder. As a result, a lightweight segmentation network with good segmentation results is developed. Song et al. \cite{song2020end} have proposed an end-to-end approach for skin lesion analysis where they jointly performed detection, segmentation, and classification using focal loss and Jaccard distance to improve the class imbalance and segmentation performance.
CMM-Net is a network proposed by Al-Masni et al. \cite{al2021cmm} for biomedical image segmentation. In the UNet design, the authors have combined the global contextual features of multiple spatial scales.

% Three datasets were used to test the proposed method: skin, retina, and brain. On all three datasets, their method outperforms others.
Khadga et al. have proposed an optimization-based few-shot approach for medical image segmentation \citep{khadga2021few, khadka2022meta}. The authors suggested employing bi-level optimization to solve the vanishing gradient problem. During training, a compound loss function consisting of log-cosh-dice and binary cross-entropy loss is used. The results of the experiments show that the proposed method performs well and has a higher generalisation ability. Zhang et al. \cite{zhang2021rich} have proposed an approach to extract rich embedding features like global embedding, peak embedding, adaptive embedding and a depth-priority context module in one-shot semantic segmentation. To incorporate the prior knowledge from support and query images, Xiao et al. \cite{xiao2021prior} proposed an approach for skin lesion segmentation for low-data regimes. The authors have extracted the prior mask from the samples (support and query) that helps discard the background and improve the segmentation performance. Sun et al. \cite{sun2021boosting} have proposed a transformer-based approach for the few-shot semantic segmentation. The authors have employed transformer blocks to extract global information and convolutional layers for local information. In continuation to this, Shen et.al. \cite{shen2021poissonseg} proposed to use semi-supervised few-shot semantic segmentation where the authors have used poisson learning to model the relationship between labelled and unlabeled samples. Also, the authors have used spatial consistency, which further improves the performance.

\subsection{Skin Lesion Classification}
Various studies have been conducted that utilizes transfer learning while classifying skin lesions \cite{elmahdy2017low, soyuslu2021new, bian2021skin}. The authors have used deep architectures and feature maps of various layers while designing the approach. These approaches demonstrate the feasibility of using deep learning in classifying skin lesions. Yu et al. \cite{yu2018melanoma} has proposed a method for detecting melanoma. The authors used a deep residual network and Fisher Vector encoding to generate a global feature representation. The authors employed SVM as a classifier, and experimental studies indicate the efficacy of their proposed method.  Huang et al. \cite{huang2019melanomanet} have proposed MelanomaNet as a melanoma detection architecture to boost feature diversity. To model the relationship between feature channels, the authors used the Inception-v4 network and included a residual-squeeze-and-excitation (RSE) block. The authors employed SVM with RBF kernel for classification. Gessert et al. \cite{gessert2019skin} have proposed an approach for addressing resolution and class-imbalance issues in dermoscopic skin lesions. The authors propose a new patch-based attention mechanism that explicitly models local and global information. To resolve the class imbalance, the authors used a loss function, diagnosis-guided loss weighting, and providing more weight to hard samples.\\
To use an ensemble approach to classify skin images, Harangi et al. \cite{harangi2018classification} have proposed the use of an ensemble consisting of deep architectures such as AlexNet, VGGNet and GoogLeNet. Tang et al. \cite{tang2020gp} used ensemble learning to develop a skin lesion classification approach that included both global and local information from the input image.
Adegun et al. \cite{adegun2020fcn} have proposed a deep-learning approach to automatically segment and classify skin lesion images. The authors have used multi-scale encoder-decoder architecture and FCN-based DenseNet for segmentation and classification. Rodrigues et al. \cite{rodrigues2020new} have proposed an IOT-based solution for the classification of skin lesions by using a web service called LINDA, which performs all the computations over the cloud. Qin et al. \cite{qin2020gan} have presented the use of style-based GAN to generate new data for training in skin lesion classification. Other approaches have posed skin lesion analysis as a few-shot problem and tried to design methods that work well with fewer data.\\
To improve the existing prototype-based method for skin lesion classification, Zhuet. al. \cite{zhu2021temperature} proposed a temperature network that generates query-specific prototypes leading to compact intra-class distributions. In this direction,  Roy et. al. \cite{chowdhury2022influential} proposed to evaluate each sample's influence (maximum mean discrepancy) with mean embedding on the sample distribution of the particular class. To further strengthen the concept of using prototypes, Prabhu et al. \cite{prabhu2019few} have proposed a prototypical clustering network for dermatological conditions. They selected more than one prototype for a class and then refined them during training. Mahajan et al. \cite{mahajan2020meta} have proposed a few-shot learning approach for the identification of skin diseases. The authors have used group equivariant convolutions (G-convolutions) instead of standard convolutions, which improves the network's performance. Li et al. \cite{li2020difficulty} have proposed a meta-learning-based approach for rare disease diagnosis. The authors have used a dynamically scaled cross-entropy loss that automatically down-weights the easy tasks and focuses on hard tasks. To use the attention mechanism for feature extraction, Liu et. al. \cite{liu2022few} proposed to use a relative position network (RPN) and relative mapping network (RMN). The authors used RPN to extract features, and RMN was used to obtain the similarity during the classification process. \\
The aforementioned algorithms yield good accuracy on the task at hand. However, they only perform segmentation or classification, and none of them integrates the two tasks in an end-to-end manner for the low data regime scenario. Also, limited algorithms incorporate explainability in few-shot learning-based algorithms for skin lesion analysis. To the best of our knowledge, this is the first paper to propose an explainable few-shot learning-based end-to-end system for skin lesion analysis that employs a segmentation mask as attention while classifying the input images.
\section{Methodology}
\begin{figure*}[t]
  \centering
  \centerline{\includegraphics[width=\textwidth]{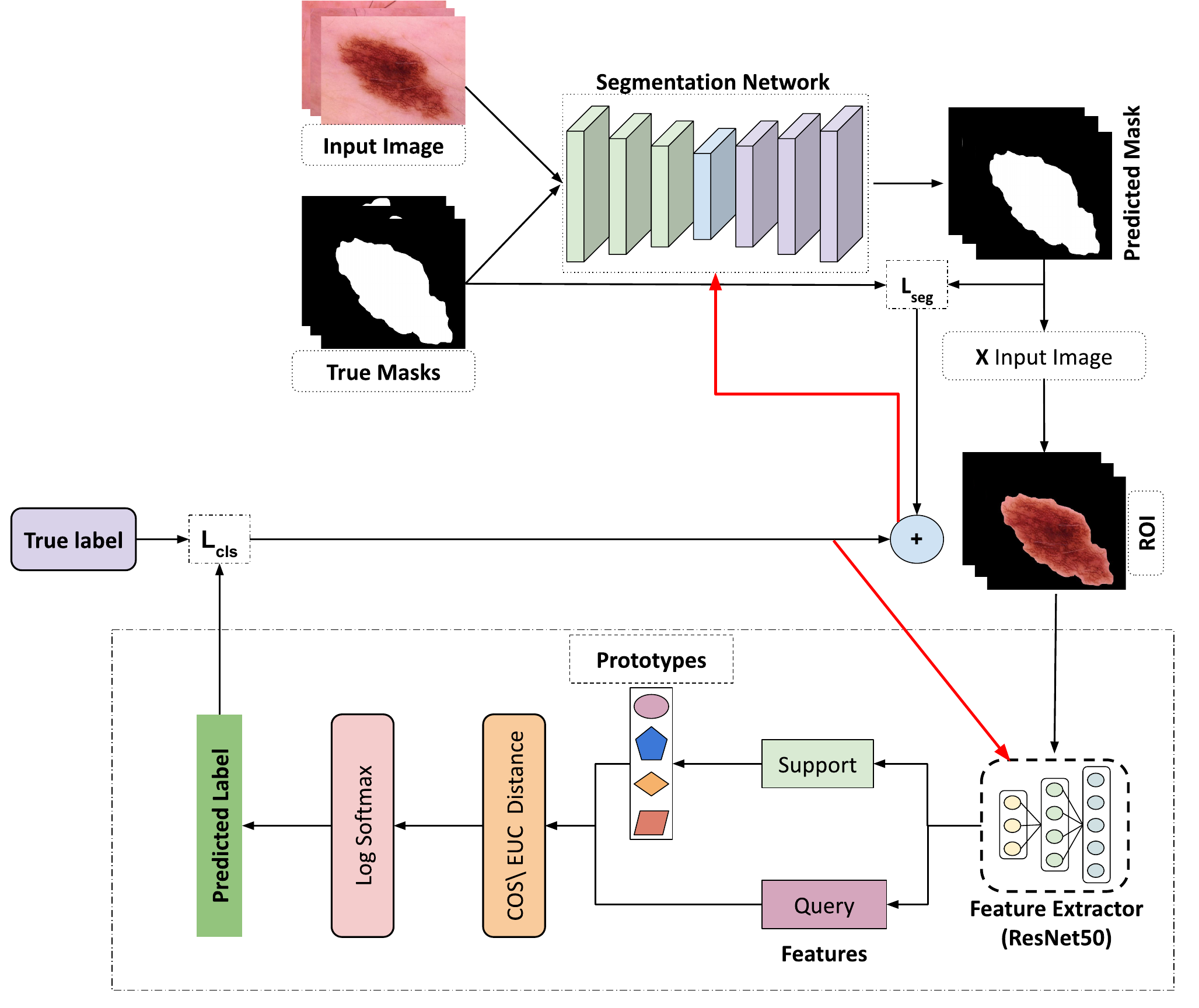}}
%  \vspace{2.0cm}
  \caption{Showcasing the proposed multitask approach that fuses the segmentation and few-shot-based classification}
  \label{fig:pipeline}
  \end{figure*}

This paper presents a multitask deep learning few-shot-based appoach for skin lesion classification. The steps involved in the proposed network are summarized in Fig. \ref{fig:pipeline}.
\subsection{Problem Definition}
The problem is to train a multi-task model that can jointly perform skin segmentation and classification on a dataset with a small number of labelled samples. We have adopted a prototypical-based approach for classification and an encoder-decoder-based architecture for segmentation. 
\subsection{Multitask Network for Segmentation and Classification}
\begin{figure*}[!ht]
  \centering
  \centerline{\includegraphics[width=0.8\textwidth]{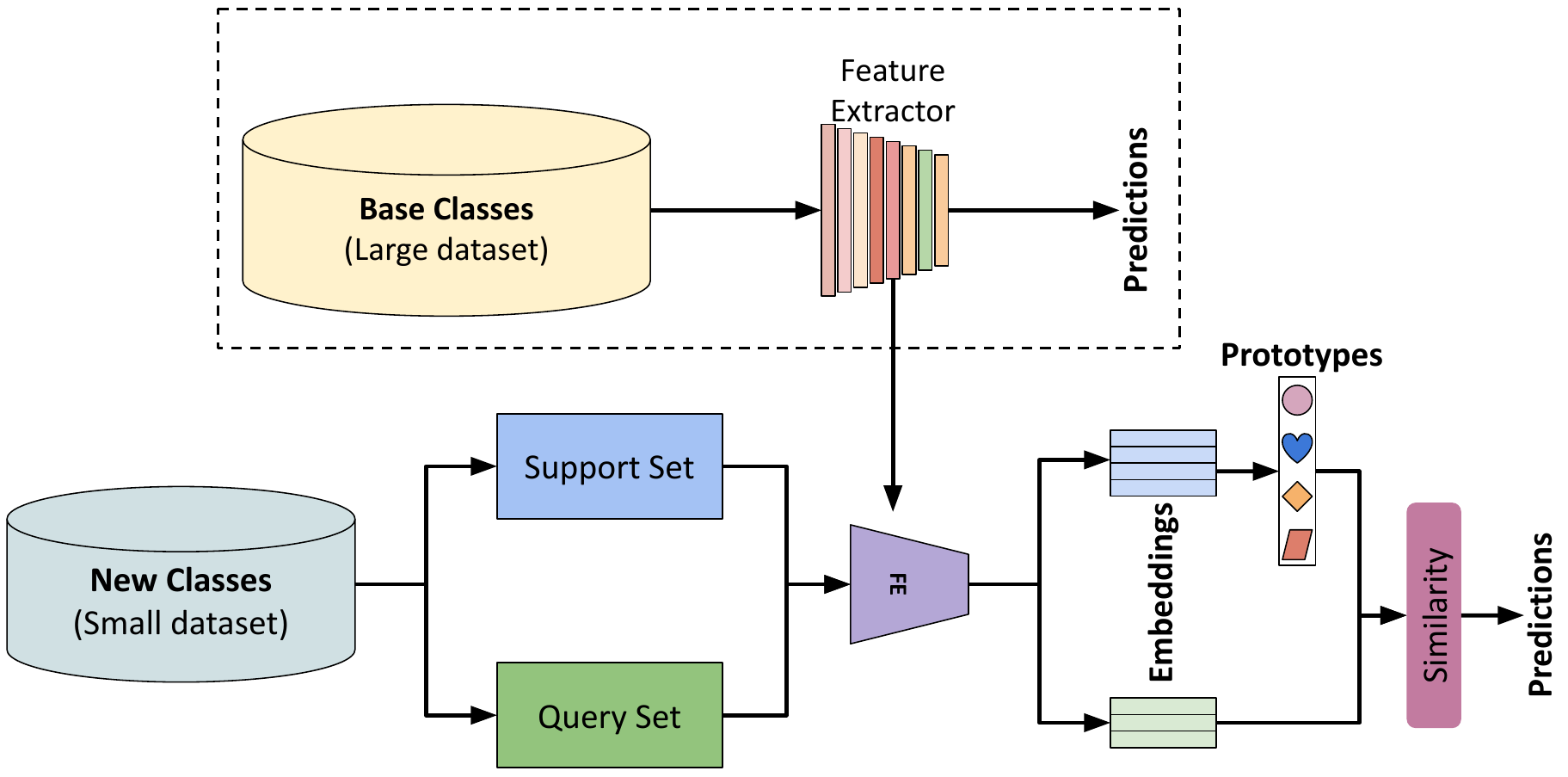}}
%  \vspace{2.0cm}
  \caption{Illustrating the basic working of prototypical networks}
  \label{fig:basics}
  \end{figure*}
This section explains the working of the proposed approach.
\subsubsection{Segmentation Network}
A segmentation model is generally an encoder-decoder-based architecture where the encoder outputs the encoding, and the decoder uses it to construct the prediction binary mask. In this work, for segmentation, we have used UNet \cite{ronneberger2015u} as it has shown significant performance in medical images \cite{fu2018joint}. Almost all skin lesion images have one region that has to be classified; hence, the segmentation model is trained using skin lesion images only. We train the segmentation network using the skin lesion images from publicly available datasets, including those used in this work. For each image $X_{i}$, segmentation loss is computed between the predicted mask and the ground truth mask using Eq. \ref{BCE}.
\begin{equation}
\label{BCE}
L_{s}=-\sum_{x, y}[\mathbf{y}_{i}(x, y) \log (\widehat{\mathbf{y}}_{i}(x, y))] \nonumber + 
(1-\mathbf{y}_{i}) \log (1-\widehat{\mathbf{y}}_{i}(x, y))
\end{equation}
where $L_{s}$ is the segmentation loss for image $X_{i}$, $y_{i}(x,y)$ and $\widehat y_{i}(x,y)$ are the pixel value of ground truth and predicted masks at point (x,y), respectively.
\subsubsection{Classification Network}
This work follows the concept of the prototypical approach for classification as in \cite{snell2017prototypical}. Fig. \ref{fig:basics} illustrates the basic working of such networks.

Similar to previous few-shot learning-based approaches \cite{ravi2016optimization,snell2017prototypical,vinyals2016matching}, we also follow an episodic way of learning. In episodic learning, each few-shot task is treated as an episode, meaning the model must learn to recognize novel classes based on only a few examples.

In prototypical networks, features are extracted from support and query images which act as the input to the rest of the pipeline. For each class in the support set, a representative vector (mean feature vector) is computed as the class prototype, and a distance is calculated between the query and the mean vectors. To predict the class label of the query image, the query image is assigned to the class with the closest distance, and a softmax activation function is used to get the class probabilities.
 Formally, the problem can be defined as:
\par Let $\mathcal{L} =\left\{l_{1},\ldots,l_{x}\right\}$ be the set of seen classes with a large number of labelled samples and $\mathcal{N} =\left\{n_{1},\ldots,n_{y}\right\}$ be the set of new and rare classes with fewer labelled samples, where $x$ and $y$ are the total number of samples in $\mathcal{L}$ and $\mathcal{N}$, respectively. The two sets are disjoint i.e., $\mathcal{L} \cap \mathcal{N}=\phi$. From $\mathcal{L}$, we can construct a large number of classification tasks by repeated sampling without replacement. Each task consists of a support set and a query set. The aim is to predict the labels of the query images based on the support images. 
\subsection{Fused Network}
This section presents the fusion of the two networks to obtain a unified pipeline.
The segmentation model helps to focus on the region of interest and is used in two experiments. 
\begin{itemize}
      \item \textbf(E1) The segmentation model is frozen (no training done) and used in the fused network to generate masks without additional refinement.
      \item \textbf(E2) We freeze some layers of the pretrained segmentation model and only train the last layer with the classification task to predict the segmentation masks. These masks are refined during back-propagation of the segmentation loss. This step helps to improve the classification accuracy.
  \end{itemize}
The overall working of the fused network is as follows: The input images and ground truth masks pass through the segmentation network. The prediction masks are generated for the images. For each predicted mask, segmentation loss is calculated using Eq. \ref{BCE}. The predicted mask is then used to extract the ROI from each respective input image using pixel-wise multiplication. These images are fed to the classification network, where they are divided into support and query images. In this work, ResNet50 \cite{he2016deep} pre-trained on ImageNet is employed as the backbone for feature encoding. Episodic training is followed to learn the embedding function by optimizing the classification loss. To compute the distance between query and class prototypes, we have used euclidean and cosine distances, which are given in Eq. \ref{ED} and \ref{CD}. 
Let Q represent the query image, and M\_i represent the mean vector of the ith class. The Euclidean distance (EUC) is calculated as follows:
\begin{equation}
\label{ED}
    EUC( Q, M\_i)   = \sqrt {\sum _{i=1}^{n}  ( Q_{i}-M_{i})^2 } 
\end{equation}
and the Cosine distance is calculated as:
\begin{equation}
\label{CD}
   CD (\theta ) =   \dfrac {Q \cdot M\_i} {\| Q\| _{2}\| M\_i\| _{2}}
\end{equation}
The classification loss is computed by taking log-softmax over the computed distances. Eq. \ref{eq:cls} gives the class probability of the query sample x and the prototype of each class $p_k$. Learning is performed by minimizing the negative log-probability of distance via the optimizer. The classification loss is computed as,
\begin{equation}
\label{eq:cls}
p_{\theta}(y=k \mid \mathbf{x})=\frac{\exp (-d(f_{\theta}(\mathbf{x}), \mathbf{p}_{k}))}{\sum_{k^{\prime}} \exp (-d(f_{\theta}(\mathbf{x}), \mathbf{p}_{k^{\prime}}))}
\end{equation}
\begin{equation}
\label{eq:losscls}
L_{c} = -\log p_{\theta}(y=k \mid \mathbf{x})
\end{equation}
The final segmentation loss that gets backpropagated is calculated by adding the segmentation loss with the classification loss given in Eq. \ref{eq:segtotal}. Here, the value of $\lambda$ is set to 2. 
\begin{equation}
\label{eq:segtotal}
L_{total} = L_{s} + \lambda \times L_{c}
\end{equation}
The classification loss is given more weightage, so the segmentation mask will be more refined. This way, the classification performance gets improved, and the same can be seen in experimental results (Tables \ref{tab:AVG_HAM}, \ref{tab:AVG_PH2}, and \ref{tab:AVG_D7}).

During testing for new and rare classes, the trained feature extractor is used for encoding, and the same episodic learning is used. The performance of the approach is evaluated using average accuracy, which is computed over the samples from the randomly chosen classes (which can be less than or equal to the number of new and unseen/rare classes).
 
\section{Experimental Setup}
 This section presents the dataset description, implementation details and experimental results for the experiments.
\subsection{Dataset Description and Preparation}
The proposed approach is evaluated on the three benchmark dermatological datasets, viz., HAM10000 \cite{tschandl2018ham10000}, PH2 \cite{mendoncca2013ph}, and Derm7pt \cite{kawahara2018seven}. These datasets are chosen because they consist of fewer samples that can act as unseen/rare classes during testing. \\
 \textbf{HAM10000} contains 10,015 dermatoscopic images corresponding to seven classes - Melanoma (1113), Melanocytic Nevi (6705), Basal Cell Carcinoma (514), Actinic Keratoses and Intraepithelial Carcinoma (327), Dermatofibroma (115), Benign Keratosis (1099), and Vascular lesions (142).\\
 \textbf{PH2} consists of a set of 200 dermoscopic images corresponding to 3 classes - Common nevi (80), Atypical nevi (80), and Melanoma (40). \\
 \textbf{Derm7pt} contains over 2000 (clinical and dermoscopy) images pertaining to 20 classes. We have used the dataset split mentioned in \cite{mahajan2020meta}.\\
\textbf{Data Preparation:} 
Out of the seven classes in HAM10000, we selected four classes with more samples (Melanoma, Benign Keratosis, Melanocytic Nevi, Basal Cell Carcinoma) in the training set. The testing set included the remaining 3 classes (Actinic Keratoses and Intraepithelial Carcinoma, Dermatofibroma, and Vascular Lesions). To verify the generalizability of the proposed approach, we used the PH2 dataset and test classes of Derm7pt only during evaluation to perform the cross-database experiments. All the images were normalized and resized to (224,224,3) to reduce the computational cost. We used several augmentation techniques on the input training images, such as vertical and horizontal flipping (with p = 0.5) and colour jittering, to add variation in the training data.
The performance of the model is evaluated using $k-$way $n-$shot learning protocols. Particularly, in our experiments, $k=$ 2 classes with varying $n=$ 1, 3 and 5. In other words, the support set comprises $k$=2 classes and 
\begin{itemize}
    \item for each class, if the number of samples is 5, it is referred to as the 5-shot setting
    \item for each class, if we have 10 samples each, it is referred to as the 10-shot setting
\end{itemize}

For comparison with the baseline, the proposed model is also evaluated on 3-way with 5-shot and 10-shot settings for the classification task.
\subsection{Implementation Details}
The proposed framework is implemented in python on the PyTorch library. The models have been trained and tested using Nvidia 1080Ti GPU. As a preliminary step for the segmentation model, we take UNet \cite{ronneberger2015u} architecture pre-trained on ImageNet \cite{deng2009imagenet} and finetuned it using publicly available skin datasets. In the fused multitask network, we trained only the last layer of the network. Eq. \ref{BCE} is used to calculate the loss between the predicted mask and the ground truth mask. For classification, we take ResNet50 \cite{he2016deep} as the backbone network for feature extraction. We used Adam as the optimizer, and the learning rate is set as 1e-3 for segmentation and 1e-6 for the classification network. We jointly train segmentation and classification networks for 10 epochs with 100 few-shot tasks each. For each experiment, all images are resized to (224x224x3) to reduce the computation.
\subsection{Experimental Results}
The performance of the proposed approach is evaluated for 3 benchmark dermatological datasets. For evaluation, we have used average accuracy (\%) as the performance metric. During the testing phase, the images are distributed equally across all the classes so that the accuracy does not get affected by the class imbalance. The reported accuracy is the average accuracy across 100 few-shot tasks sampled from the test set. The model is evaluated by varying $k$ as 1, 3 and 5, and the distance metric as cosine and euclidean distance for the 2-way classification task.\\
%To the best of our knowledge, there is no approach that explored HAM10000 and PH2 in the few-shot learning setting as we proposed in this work. 
Comparative results for various models on HAM10000, PH2 and Derm7pt are summarized in Tables \ref{tab:AVG_HAM}, \ref{tab:AVG_PH2}, \ref{tab:AVG_D7} respectively.
The average accuracy for all the datasets goes on increasing as the value of $k$ increases. This can be attributed to the fact that more samples provide more information to represent the underlying distribution and hence can estimate the class prototype in a better way. For skin lesions, it is observed that the proposed method outperforms other methods on all three datasets. In case of HAM10000, the average accuracy for 1, 3 and 5 shots get improved by around 2\%; in PH2, there is a significant improvement of average accuracy from 70.98\% (ProtoNet with Cosine distance) to 76.06\% (Proposed Network with Euclidean distance) for $k$ = 5; for $k$ =1 and 3, the proposed network also outperforms the other networks. For the Derm7pt dataset, the proposed approach is also able to outperform the other methods. It is quite evident from the experimental results that both cosine distance and euclidean distance are effective distance metrics for few-shot prototypical-based classification networks in skin lesion classification.

\begin{figure}[!ht]
  \centering
  \includegraphics[width=\columnwidth]{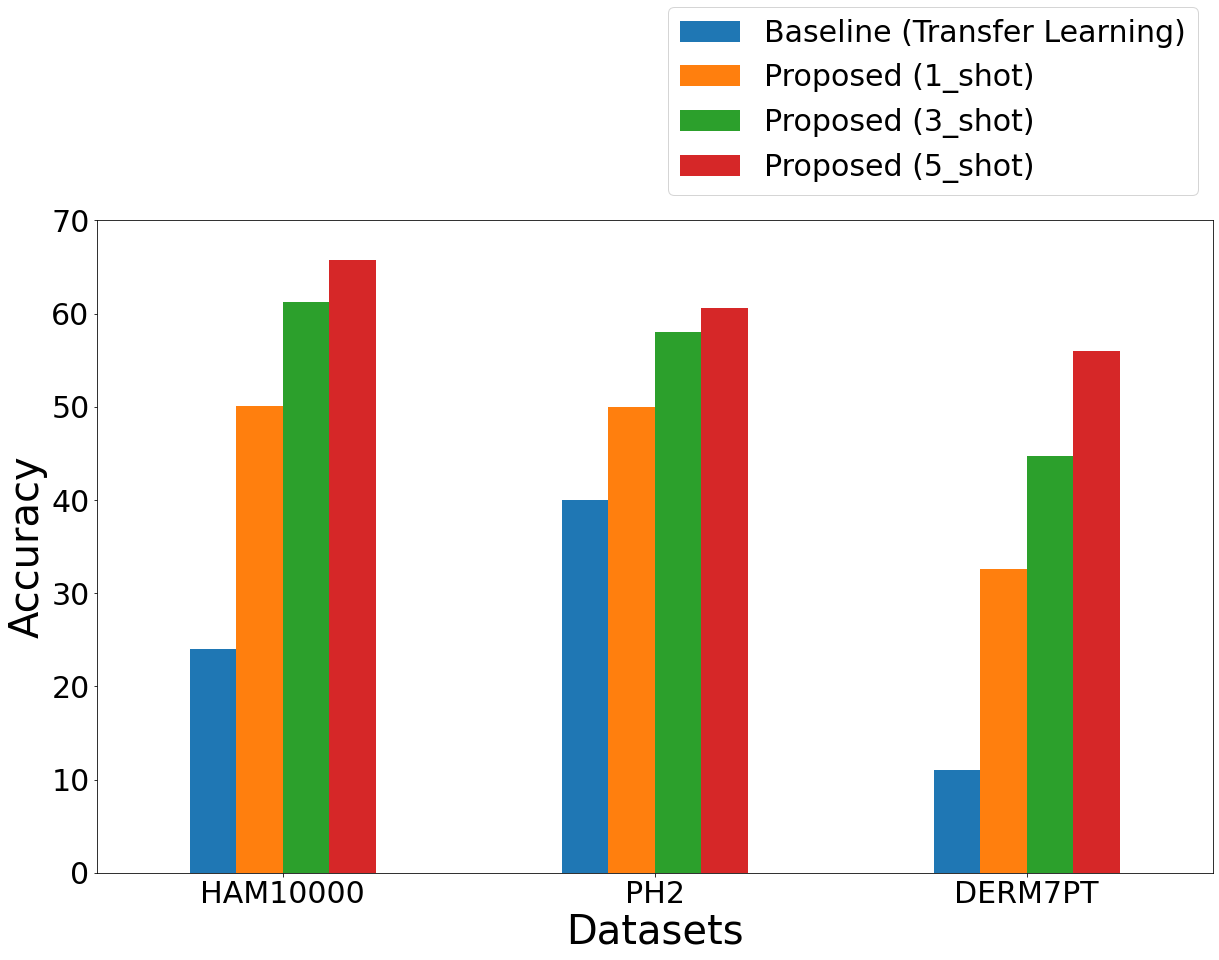}
  \caption{Comparing the performance of the proposed approach (with cosine distance) with baseline (transfer learning) for 3-Way and 5-way classification. Each set of bars corresponds to the average accuracy on HAM10000, PH2 and Derm7pt, respectively.}
\label{fig:compa}
  \end{figure}

As a baseline experiment, a pre-trained ResNet50 model is trained on the training data (4 classes from HAM10000, same as used in the proposed approach). All the parameters, such as learning rate, epochs, and optimizer, are kept the same for a fair comparison. For testing, the model is fine-tuned for the rare classes (same as in the proposed approach). The average accuracy for the test classes of HAM10000, PH2 and Derm7pt is shown in Fig. \ref{fig:compa}. This accuracy is low compared to the accuracy obtained by the proposed approach on 3-way classification. This behaviour is quite explainable as the pre-trained model is trained on classes different than classes at test time. The trained model is biased towards the training data and does not adapt to the new unseen data. On the other hand, the proposed approach is able to generalize well for the unseen classes as it tries to learn the metric space that helps to better classify the images. This demonstrates the limitation of transfer learning of deep models to adapt to new unseen data. We have also reported the results in terms of confidence intervals. Table \ref{tab:CI} summarizes the results for the datasets by varying the values of CI as 70, 90 and 95\%. The margin error decreases with the number of shots (samples). The margin error is used to measure the confidence of the model in its predictions. The larger margin error indicates lower confidence and vice versa. As the number of shots increases, the model gets more confident about the predictions; hence the margin of error is low. In the HAM10000 dataset, the margin error decreases from 0.51 to 0.43, 1.06 to 0.73 for 75\% and 95\% CI, respectively, indicating the model is getting more confident about the predictions. Generally, as we increase the CI, the width of CI will also increase and result in a higher margin of error. However, other factors, such as data and sample size variability, also affect the margin of error and can be seen in the behaviour of 90\% CI for the HAM10000 dataset. In future, we will try to build a more resilient and robust system that will have a lower margin of error.
\begin{table}[!ht]
\caption{Comparison of Average Accuracy (in \%) on HAM10000 dataset for 2-way classification (shots = 1,3,5)}
\centering
\resizebox{\columnwidth}{!}{%
\begin{tabular}{lcccc}
\hline
\multicolumn{1}{c}{Model}                                    & Distance  & 1              & 3              & 5              \\ \hline
Relation Net \cite{sung2018learning}                                                & -         & 49.65          & 50.5           & 56.35          \\ 
Matching Net \cite{vinyals2016matching}                                                & -         & 65.9           & 73.3           & 75.7           \\ 
\begin{tabular}[c]{@{}l@{}} MDD \cite{mahajan2020meta}  \end{tabular} & -    & 64.5   & 73.5  & \textbf{79.7}\\
\begin{tabular}[c]{@{}l@{}}ProtoNet \cite{snell2017prototypical}\end{tabular}      & Cosine    & 64.47          & 72.91          & 77.71          \\ 
\begin{tabular}[c]{@{}l@{}}ProtoNet \cite{snell2017prototypical}\end{tabular}      & Euclidean & 60.22          & 70.04          & 76.35          \\
E1                                               & Cosine        &   62.39         &  71.72        &     75.17      \\ 
\textbf{\begin{tabular}[c]{@{}l@{}}Proposed \end{tabular}} & Euclidean & \textbf{62.65} & 73.12          & 77.57          \\ 
\textbf{\begin{tabular}[c]{@{}l@{}}Proposed \end{tabular}} & Cosine    & 62.24          & \textbf{73.85} & \textbf{78.21} \\ \hline

\end{tabular}
}
\label{tab:AVG_HAM}
\end{table}
\begin{table}[!ht]
\caption{Comparison of Average Accuracy (in \%) on Derm7pt dataset for 2-way classification (shots = 1,3,5)}
\centering
\resizebox{\columnwidth}{!}{%
\begin{tabular}{lcccc}
\hline
\multicolumn{1}{c}{Model}                                    & Distance  & 1              & 3              & 5              \\ \hline
Relation Net \cite{sung2018learning}                                                & -         & 50.35          & 50.25         & 49.65        \\ 
Matching Net \cite{vinyals2016matching}                                                & -         & 59.30           & 66.55         & 65.60        \\ 
\begin{tabular}[c]{@{}l@{}} MDD \cite{mahajan2020meta}  \end{tabular} & -    & 64.1   & 66.8  & 69.5\\

SCAN \cite{li2022sub} & - & \textbf{65.27} & - &\textbf{ 77.97}\\

\begin{tabular}[c]{@{}l@{}}ProtoNet \cite{snell2017prototypical}\end{tabular}      & Cosine    & 60.68         & 70.74        & 76.54       \\ 
\begin{tabular}[c]{@{}l@{}}ProtoNet \cite{snell2017prototypical}\end{tabular}      & Euclidean & 60.62        & 70.80        & 69.91        \\
E1                                               & Cosine        &   59.48         &   59.95       &   59.95        \\ 
\textbf{\begin{tabular}[c]{@{}l@{}}Proposed \end{tabular}} & Euclidean & 59.95 & \textbf{71.71}         & 77.64\\
\textbf{\begin{tabular}[c]{@{}l@{}}Proposed \end{tabular}} & Cosine    & 62.70      & \textbf{71.71} & \textbf{77.83} \\
 \hline

\end{tabular}
}
\label{tab:AVG_D7}
\end{table}

\begin{table}[!ht]
\caption{Comparison of Average Accuracy (in \%) on PH2 dataset for 2-way classification (shots = 1,3,5)}
\centering
\resizebox{\columnwidth}{!}{%
\begin{tabular}{lcccc}
\hline
\multicolumn{1}{c}{Model}                                    & Distance  & 1              & 3              & 5              \\ \hline
Relation Net \cite{sung2018learning}                                                & -         & 50.45          & 49.60          & 51.10         \\ 
Matching Net \cite{vinyals2016matching}                                                & -         & 59.75           & 65.0          & 64.05         \\ 
\begin{tabular}[c]{@{}l@{}}ProtoNet \cite{snell2017prototypical}\end{tabular}      & Cosine    & 61.59         & 68.88         & 70.98         \\ 
\begin{tabular}[c]{@{}l@{}}ProtoNet \cite{snell2017prototypical}\end{tabular}      & Euclidean & 57.57         & 66.82        & 69.91         \\ 
E1                                               & Cosine        &   61.31         &   68.39       &    71.86       \\ 
\textbf{\begin{tabular}[c]{@{}l@{}}Proposed \end{tabular}} & Euclidean & 59.86 & 68.89          & \textbf{76.06}         \\ 
\textbf{\begin{tabular}[c]{@{}l@{}}Proposed \end{tabular}} & Cosine    & \textbf{63.65}        & \textbf{70.81} & 75.85 \\ \hline

\end{tabular}
}
\label{tab:AVG_PH2}
\end{table}
\subsection{Visualization results}
\begin{figure}[ht]
  \centering
  \includegraphics[width=\columnwidth]{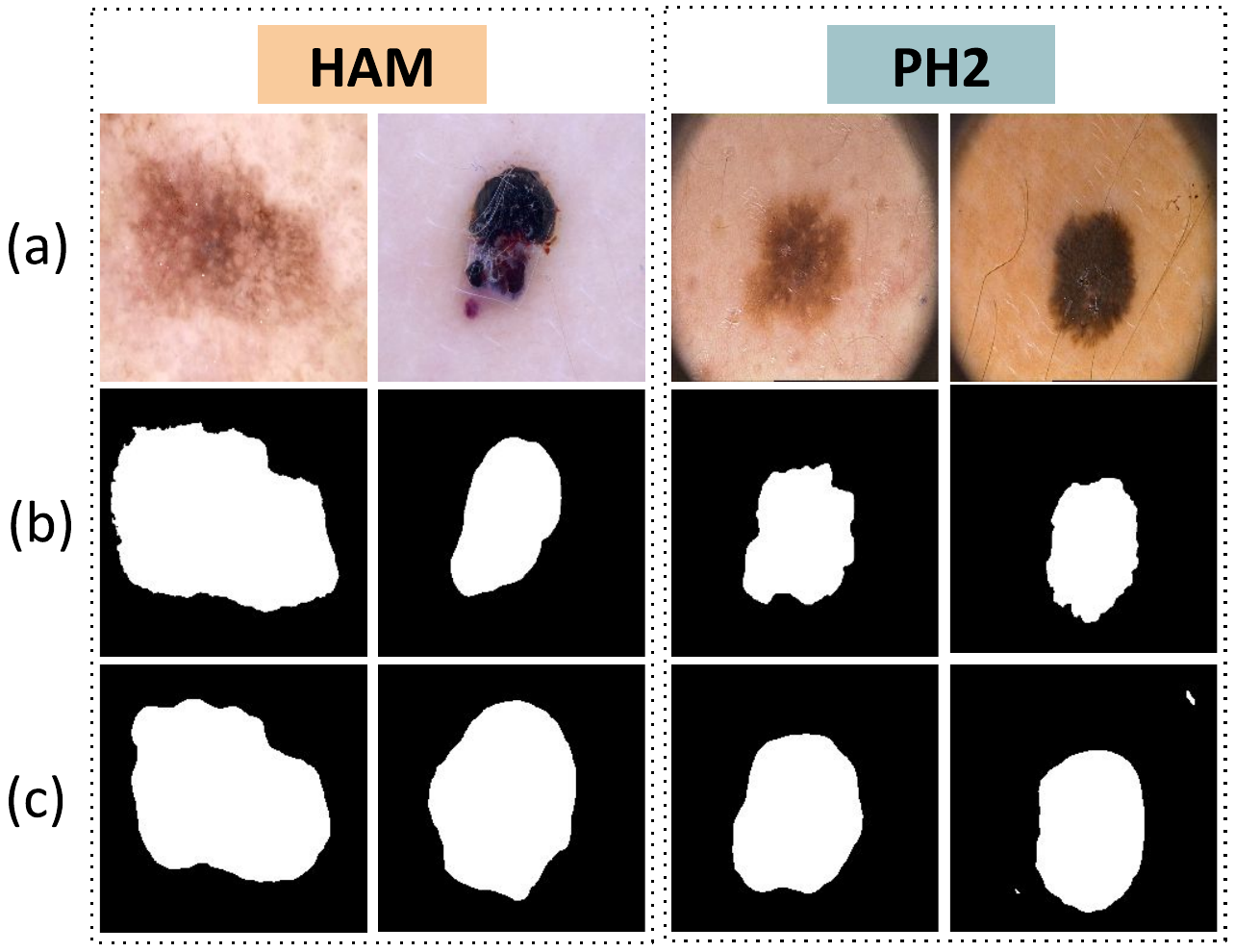}
%  \vspace{2.0cm}
\caption{Showing the images (a), their ground truth masks (b) and the predicted masks (c) by the segmentation network while training the last layer of it in parallel to the classification network.}
\label{fig:seg_results}
  \end{figure}

\begin{figure*}
  \centering
  \includegraphics[width=\textwidth]{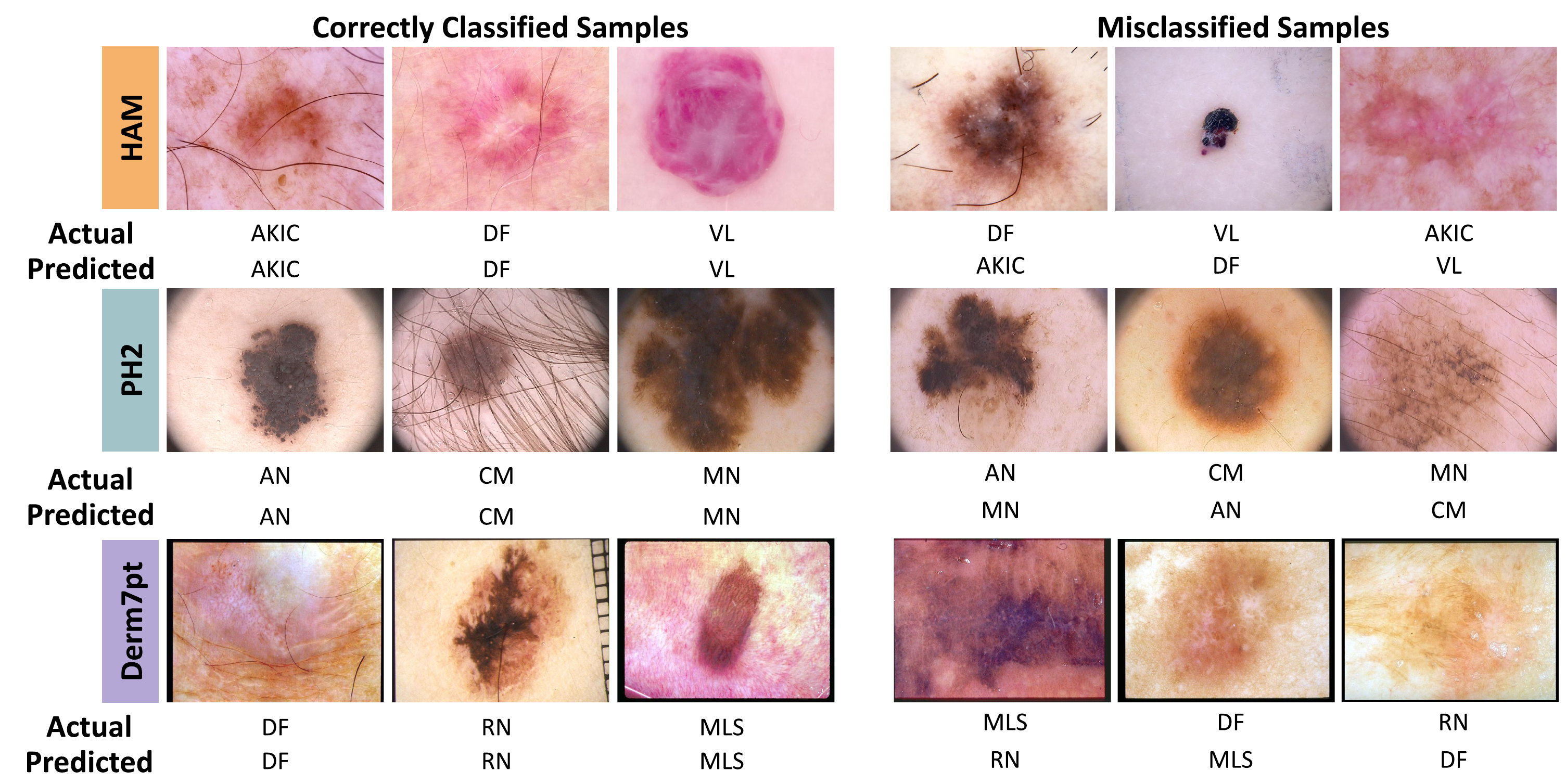}
%  \vspace{2.0cm}
\caption{Showcasing some of the correctly and misclassified samples for the three datasets.}
\label{fig:visual}
  \end{figure*}

We have reported some segmentation visualisations in Fig. \ref{fig:seg_results}. The proposed algorithm can produce a segmentation mask that is quite similar to the ground truth mask. This refinement of masks, even with fewer samples, can be attributed to utilising classification loss in the segmentation network. This segmentation phase enables the algorithm to avoid the artefacts in the image and focus just on the affected area. However, as shown in Fig. \ref{fig:visual}, images with fading borders and low contrast require closer attention. 

\begin{figure*}[ht]
  \centering
  \includegraphics[width=\textwidth]{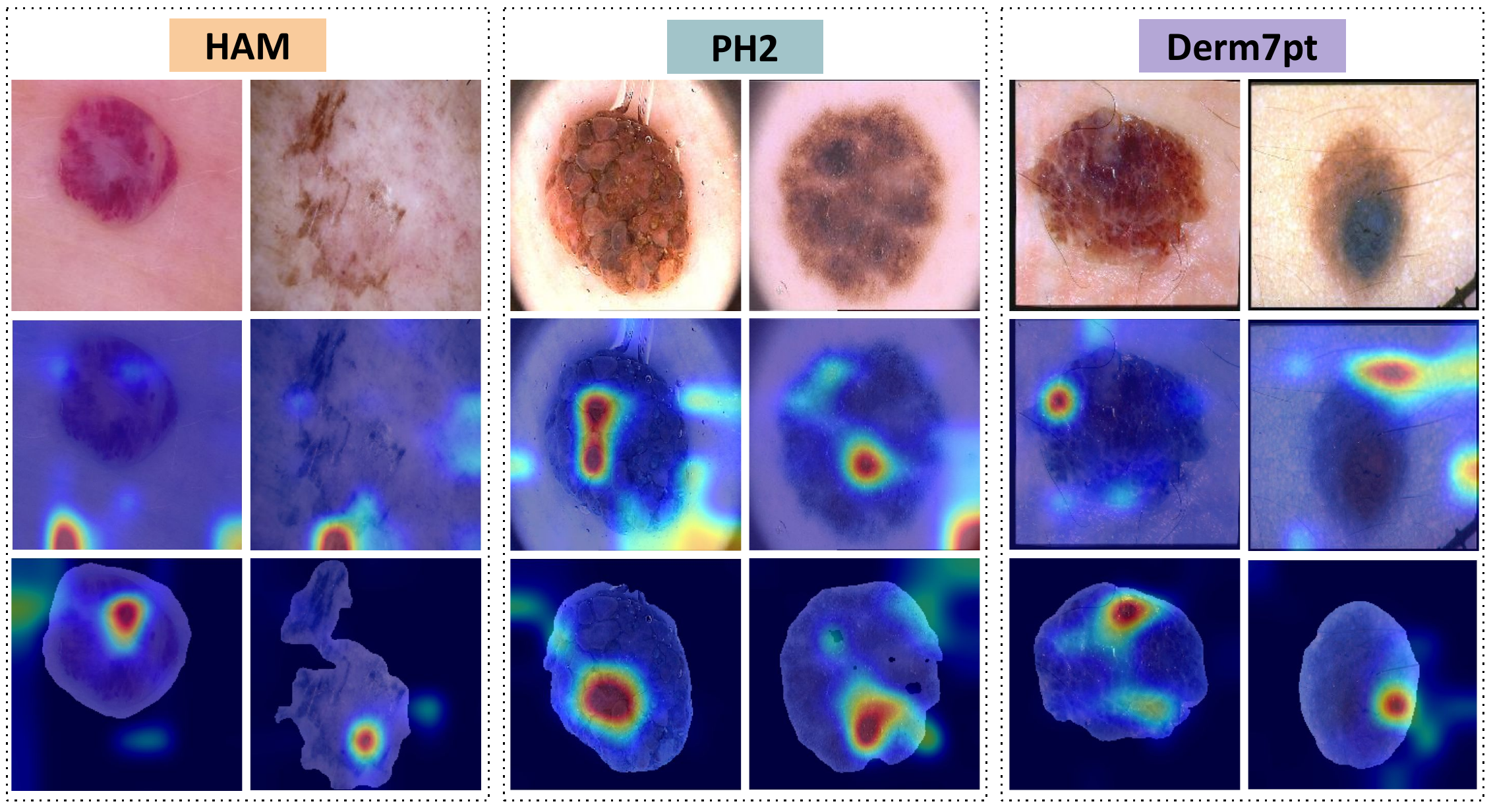}
%  \vspace{2.0cm}
\caption{Visualization of the GradCAM results on the samples of the three datasets. The top row represents the input image, the middle row represents the GradCAM results without segmentation, and the bottom row represents the GradCAM results of the proposed approach. The red areas represent the regions that were taken into account while making a decision (Best viewed in color).}
\label{fig:grad_results}
  \end{figure*}

To further support the use of the segmentation phase, we have also used GradCAM \cite{selvaraju2017grad} to showcase the visual explanations. By highlighting the areas on which the algorithm focuses while making a decision, it aids in explaining the decisions made by the proposed approach. As observed in Fig. \ref{fig:grad_results}, the proposed approach focuses only on the affected region. However, without segmentation, the algorithm examines a variety of unaffected visual components and, therefore, of little interest. This can also be seen in Fig. \ref{fig:grad_results}. Since there are fewer test samples, the proposed approach focuses on the smallest region that discriminates the samples in a better way. This explains the smaller portion being highlighted in the GradCAM results. These explanations help the end-users to get a better idea of the working of the algorithm and increase the credibility of any AI algorithm.

\begin{table}[!ht]
\centering
\caption{Comparison of Average Accuracy (in \%) over 1000 randomly generated episodes with varying confidence interval (75, 90 and 95) on HAM10000 (HAM), Ph2, and Derm7pt dataset for 2-way classification (shots = 1,3,5)}
\resizebox{\columnwidth}{!}{%
\begin{tabular}{ccccc}
\hline
\multicolumn{2}{c}{\begin{tabular}[c]{@{}c@{}}CI/Datasets\end{tabular}} & \textbf{75}          & \textbf{90}          & \textbf{95}          \\ \hline
\multirow{3}{*}{\textbf{HAM}}                            & 1                        & 62.65 $\pm$ 0.51 & 62.65 $\pm$ 0.55 & 62.65 $\pm$ 1.06 \\
 & 3                        & 73.12 $\pm$ 0.47 & 73.12 $\pm$ 0.67 & 73.12 $\pm$ 0.80 \\
                                                & 5                        & \textbf{77.57}  $\pm$ \textbf{0.43} & \textbf{77.57}  $\pm$\textbf{0.61} & \textbf{77.57} $\pm$ \textbf{0.73} \\ \cline{1-1} \hline
\multirow{3}{*}{\textbf{Ph2}}                            & 1                        & 59.86$\pm$ 0.62 & 59.86 $\pm$ 0.89 & 59.86  $\pm$ 1.06 \\
                                                & 3                        & 68.89 $\pm$ 0.59 & 68.89 $\pm$ 0.84 & 68.89 $\pm$ 1    \\
                                                & 5                        & \textbf{76.06} $\pm$ \textbf{0.57} & \textbf{76.06} $\pm$ \textbf{0.82} & \textbf{76.06} $\pm$  \textbf{0.98 }\\ \cline{1-1} \hline
\multirow{3}{*}{\textbf{Derm7pt}}                        & 1                        & 59.95 $\pm$ 0.62 & 59.95 $\pm$ 0.89 & 59.95 $\pm$ 1.06 \\
                                                & 3                        & 71.72 $\pm$ 0.6  & 71.72 $\pm$  0.85 & 71.72 $\pm$ 1.02 \\
                                                & 5                        & \textbf{77.64} $\pm$ \textbf{0.56 }& \textbf{77.64} $\pm$ \textbf{0.80} & \textbf{77.64}  $\pm$ \textbf{0.96 }\\ \cline{1-1} \hline
\end{tabular}%
}
\label{tab:CI}
\end{table}
  
\subsection{Ablation Study}
To study and investigate the effect of varying feature extractors and the value of $\lambda$, we have performed various experiments to showcase the effect on classification accuracy.
\subsubsection{Effect of Feature Extractor}
By varying the backbone for feature extraction with fewer layers (VGG16, ResNet18) and a large number of layers (DenseNet-121), the proposed algorithm is evaluated. Table \ref{tab:FE} presents the reported results for the 5-shot 2-way setting. The results are clearly inferior to those shown in Tables \ref{tab:AVG_HAM}, \ref{tab:AVG_PH2} and \ref{tab:AVG_D7}. The smaller sample size in each rare class can be attributable to this pattern. The prototypical networks are based on the calculation of a mean /representative vector with which the distances are computed. Choosing the right architecture helps to make the approach more efficient. Less layered architecture does not allow for the selection of more distinctive features, and more layered architecture reduces generalizability due to overfitting. Based on the results, we opted to use ResNet50 as the backbone for feature extraction in this work.

\subsubsection{Effect of $\lambda$ in loss function}
We evaluated the proposed algorithm by changing the value of $\lambda$ in \ref{eq:segtotal} in order to examine the impact of combining the classification loss and segmentation loss. The results are reported in Table \ref{tab:lamb}. 
\begin{table}[!ht]
\caption{Average Accuracy (in \%) by varying feature extractor for 5-shot 2-way classification}
\centering
\begin{tabular}{lccc}
\hline
\multicolumn{1}{c}{Feature Extractor}                                    &  HAM10000            & PH2             &  Derm7pt             \\ \hline
VGG16 \cite{SimonyanZ14a}                                                  & 73.79          & 66.33        & 68.37       \\ 
ResNet18 \cite{he2016deep}                                                       & 76.50          & 74.35         & 75.05      \\ 
DenseNet-121 \cite{huang2017densely}                               & 74.21           & 73.95        &    72.23     
\\ 
\hline
\end{tabular}
\label{tab:FE}
\end{table}
As we can see from the results, combining the classification loss with the segmentation loss improves the classification accuracy. The misclassification directs the segmentation network to refine the mask, which acts as attention for the classification network to focus only on the affected segmented region. This step helps in improving both classification and segmentation performance. The accuracy, however, declines if we increase the weight of classification loss or the value of $\lambda$. It may be because fewer samples are available, and giving classification loss more weight results in poor generalizability.

\begin{table}
\centering
\caption{Average Accuracy (in \%) by varying the value of $\lambda$ for 5-shot 2-way classification}

\begin{tabular}{lccc}
\hline
\multicolumn{1}{c}{ $\lambda$ }                                    &  HAM10000            & PH2             &  Derm7pt             \\ \hline

1  &     77.50      &    74.79    &    76.30      \\ 
2  &    78.21       &     75.85    &   77.83      \\ 
3  &     77.37      &    75.02     &   76.45      \\ 
4  &     77.25      &     74.89    &    76.32     \\ 

\hline
\end{tabular}
\label{tab:lamb}
\end{table}

\section{Conclusion}
Skin cancer is one of the most serious skin diseases. Skin lesions that are detected early can help prevent complications, including death. Because the trained model becomes biased towards the classes seen during training, the transfer learning approach does not generalise well to new and unseen classes. This behaviour can also be observed in medical settings, where new classes are introduced with a small sample size. This paper presents a multitask few-shot learning-based network for skin lesion classification. In the proposed approach, the segmentation and classification tasks are fused into a single pipeline. We evaluated the proposed network using Cosine and Euclidean distances, and it was proved from the experimental results that both distances are effective in the case of classifying skin lesions. Experiments with the HAM10000, PH2 and Derm7pt datasets show that the proposed method yields good accuracies even when applied across databases, exhibiting generalizability. Any skin disease can benefit from the proposed approach. To further improve the performance in future, we plan to propose a more robust segmentation network with varying loss functions. Moreover, we intend to expand our research to other medical modalities with limited data in the future.

 \bibliographystyle{elsarticle-num-names} 
 \bibliography{sample}

% %% else use the following coding to input the bibitems directly in the
% %% TeX file.




% \end{thebibliography}
\end{document}